# Conjure Revisited:
# Towards Automated Constraint Modelling

Ozgur Akgun[1], Alan M. Frisch[2], Brahim Hnich[3], Chris Jefferson[1], Ian Miguel[1]

[1]School of Computer Science, University of St Andrews, UK
[2]Artificial Intelligence Group, Dept. of Computer Science, University of York, UK
[3]Department of Computer Engineering, Izmir University of Economics, Turkey

**Abstract.** Automating the constraint modelling process is one of the key challenges facing the constraints field, and one of the principal obstacles preventing widespread adoption of constraint solving. This paper focuses on the refinement-based approach to automated modelling, where a user specifies a problem in an abstract constraint specification language and it is then automatically *refined* into a constraint model. In particular, we revisit the Conjure system that first appeared in prototype form in 2005 and present a new implementation with a much greater coverage of the specification language Essence.

## 1 Introduction

Automating the constraint modelling process (the *modelling bottleneck*) is one of the key challenges facing the constraints field [16], and one of the principal obstacles preventing widespread adoption of constraint solving. Without help, it is very difficult for a novice user to formulate an effective (or even correct) model of a given problem. This challenge has received a considerable amount of attention in the literature recently, where a variety of approaches have been taken to automate aspects of constraint modelling, including: machine learning [1]; case-based reasoning [10]; theorem proving [2]; automated transformation of medium-level solver-independent constraint models [17,15,20,13]; and refinement of abstract constraint specifications [4] in languages such as ESRA [3], Essence [5], F [7] or Zinc [11].

The refinement-based approach is the focus of this paper. The central idea is to allow a user to write *abstract* constraint specifications that describe a problem above the level at which modelling decisions are made. Abstract constraint specification languages, such as Essence or Zinc support abstract decision variables with types such as set, multiset, function, and relation, as well as *nested* types, such as set of sets, multiset of function variables, and so on. Problems can typically be specified very concisely in this way, as demonstrated by the example in Figure 1. However, since existing constraint solvers do not support these abstract decision variables directly, abstract constraint specifications must be *refined* into concrete constraint models.

The Conjure system was introduced in prototype form by Frisch *et al.* [4]. It was able to refine a fragment of Essence limited to nested set and multiset based

```
given      item_count,capacity : int
letting    item be domain int(1..item_count)
given      volume,value : function (total) item -> int(1..)
find       x            : set of item
maximising sum i elem x . value(i)
such that  (sum i elem x . volume(i)) <= capacity
```

**Fig. 1.** The knapsack problem, given in Essence

decision variables into models in the Essence′ solver-independent modelling language. This work was further developed by Frisch and Martinez-Hernandez [12], who looked in depth at the issues involved in channeling efficiently between many different representations of high-level variables.

This paper describes a new approach to the implementation of Conjure, with a much greater coverage of the Essence language and fewer limitations on the rewrite rules such as the requirement of the IJCAI prototype to flatten an Essence specification before applying the rewrite rules.

## 2  Background

Essence is a language for specifying combinatorial (decision or optimisation) problems (for a detailed description see [5]). It is motivated by a desire to allow users to write down problems *without* making constraint modelling decisions. To this end, it has a high level of abstraction, supporting decision variables whose types match the combinatorial objects problems typically ask us to find, such as: sets, multisets, functions, relations and partitions. The key advance represented by the language is its support for the *nesting* of these types, allowing decision variables of type set of sets, multiset of sets of functions *etc*. Hence, problems such as the Social Golfers Problem [6], which is naturally conceived of as finding a set of partitions of golfers subject to some constraints, can be specified directly without the need to model the sets or partitions as matrices.

Concomitant with the decision to support abstract constraint specifications is the requirement to transform these specifications into constraint models. Today's constraint solvers typically support decision variables with atomic types, such as integer or Boolean, have limited support for more complex types like sets or multisets, and no support for nested complex types. Hence, abstract specifications must be *refined* into constraint models suitable for constraint solvers. This is achieved by modelling abstract decision variables as constrained collections of variables of more primitive type.

The Conjure system, first presented in prototype form in [4], employs a system of rewrite rules to refine Essence specifications into constraint models in Essence′ [17], a language derived from Essence mainly by removing facilities for abstraction and adding facilities common to existing constraint solvers and toolkits. From Essence′ a tool such as Tailor [17] can be used to translate the model into the format required for a particular constraint solver. The new version of CONJURE presented herein operates in the same way. One could envisage an alternative mode of operation where the user writes abstract specifications, which are then refined and solved, and then the solution(s) are mapped back to the abstract level. This is the subject of future research.

There are typically many constraint models for a given abstract specification. CONJURE is intended to generate these alternatives by providing multiple refinement rules for each abstract type, corresponding to the various ways in which a decision variable of that type can be modelled. Furthermore, for each way of modelling the decision variables there can be multiple rules to generate alternative models for a constraint on those variables. Consequently CONJURE often generates a large set of alternative models for an input specification. At this stage our focus is on generating as large a set of model as possible. In future, we will investigate restricting this set to *good* models and the selection of either one recommended model or a portfolio of models with complementary strengths.

The previous CONJURE prototype provided alternative rules for the refinement of arbitrarily-nested sets and multisets. The new implementation presented in this paper extends these rules to cover relations and functions. The remainder of this paper describes the new implementation of CONJURE, first by describing its architecture and then turning attention to its refinement rules and its implementation.

## 3 The Architecture of CONJURE

CONJURE is a compiler-like system. Like most compilers it has a pipeline, which starts with parsing, validating the input, and type-checking. After these foundation phases, it prepares the input specification for rewriting, performs rewriting, and does some housekeeping. The pipeline is summarised below:

1. Parsing
2. Validating the input
   - (*all the identifiers are defined in the declarations part*)
   - (*properties of declarations. ex: a function cannot be declared both total and partial.*)
3. Type checking the input
4. Representations phase
5. Auto-Channelling phase
6. Adding structural constraints
7. Expression rewriting
8. Fixing auxiliary and quantified variable names

Phases 1–3 are the foundation phases. The representations, auto-channelling, and adding structural constraint phases (4–6) prepare the input specification for the actual task of rewriting (Phase 7). Phase 8 can be viewed as housekeeping: it makes the output models easier to read and understand. Phase 7 (expression rewriting) is described in detail in the following sections. There follow brief descriptions for the three preparatory phases preceding it.

**Representation** There are two fundamental decisions to be made in formulating a constraint model. The first is as to the *viewpoint* (the choice of decision variables and their domains [9]), and the second is the choice of constraints on those decision variables. The Conjure prototype interleaved these two decisions in a constraint-wise refinement process [4]. By contrast, the version of Conjure described herein explores all possible viewpoints first before constraint refinement. This phase takes an Essence specification and generates a *set* of specifications, one per possible combination of variable representation decisions.

If a variable appears in more than one constraint, it is possible for that variable to be represented in different ways in different constraints, in case the most efficient way of representing a complex decision variable differs among the constraints it appears in. If a variable is represented in more than one way in a specification, channelling constraints between the two (or more) representations are added automatically in Phase 5.

**Auto-Channelling** If we choose more than one way of representing a combinatorial object within a specification, we automatically add channelling constraints [12] between different representations of the same variable in this phase to maintain consistency.

**Adding structural constraints** In this phase we add all necessary *structural* constraints for every decision variable in the specification. The structural constraint for a representation of a decision variable makes sure the selected representation actually represents a valid combinatorial object with the intended properties (e.g. ensuring that the elements of a set are distinct). We add these constraints before rewriting because they will be added regardless of the rest of the specification and they only depend on the representation of a combinatorial object.

## 4 Non-deterministic Rewriting

We employ a term rewriting system to refine Essence specifications into the target language Essence'. Generally, rewrite rules can be thought of as partial functions, which map from a subterm to an *equivalent* subterm [14]. Given a set of rewrite rules and a term, a rewrite system repeatedly applies the rules until no further rules can be applied. The term is then said to be in *normal form*.

In order to produce alternative models we wish to generate not a single normal-form term, but all normal-form terms attainable by applying the given rules to the input term. Hence, we adjust the definition of a rewrite rule: instead of a function that maps from a subterm to an equivalent subterm, we define a rewrite rule to be a function that maps from a subterm to a *set* of subterms.

A single rule is now sufficient to represent the whole rule database. This single rule is a one-to-many mapping from a subterm to the set of rewrites of that subterm. This representation is natural while applying the rules, but it is not a natural way to write them. It is, however, trivial to automate the combination of a set of partial functions into the single function used by the implementation.

For example we can combine `rule1`, `rule2` and `rule3` in `allRules` as follows:

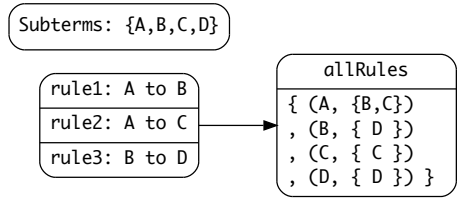

Here `rule1`, `rule2`, and `rule3` are partial functions. However, the combined `allRules` is a total function, mapping from a subterm to a set of subterms. `rule1` rewrites `A` into `B`, `rule2` rewrites `A` into `C` and `rule3` rewrites `B` into `D`. No rule matches `C` or `D`, so they are each mapped to a singleton set containing themselves.

Henceforth, we will present our rules as partial functions mapping from single subterms to single subterms. As above, these rules are then automatically combined into a single rule from subterms to sets of subterms.

Fig. 2 presents the elements of a rule: the mapping denoted by the $\rightsquigarrow$ operator; the guards that the left hand side of the mapping must satisfy; and the declarations used to construct the right hand side of the mapping. Any expression that matches the left hand side of the $\rightsquigarrow$ symbol is replaced by the right hand side, if all guards are satisfied.

In presenting rewrite rules we adopt the following convention, which is important to remember throughout reading this paper. Whenever a rule uses a guard to restrict the type of a meta-variable on the left-hand-side of $\rightsquigarrow$ that meta-variable is only allowed to match an atomic expression or an expression of the form $Array[i_1, \ldots, i_n]$ (we employ a *lifting* mechanism to handle indexed expressions as if they were just simple atomic expressions). As an example, consider the rule in Fig. 3. Here the meta-variables `a` and `b` are each guarded to match only expressions of type `set of` $\tau$, where $\tau$ is any type. For example, `a` and `b` can match against $\{1, 2, 3\}$, a decision variable or parameter whose type is `set of set of int` or `x[i]` where $x$ is a decision variable of type `matrix of set of int`. However, `a` and `b` cannot match against the expression `s1 union s2`.

Fig. 3 shows a rule that matches with a `subseteq` constraint between two sets of the same type. It rewrites the constraint into a universal quantification over the first set, which can be read as *every element in set a must also be an element of set b*. It also creates a quantified variable of type $\tau$, which is the type of the elements of the two sets.

The symbols that we can match on the left-hand-side of a $\rightsquigarrow$ are ESSENCE expression constructors. These symbols can be either data constructors or operators on expressions. Currently implemented data constructors are `bool`, `int`,

```
essence_expression ⇝ equivalent_expression
    guards: properties that essence_expression must satisfy
    declarations: newly created variables and local aliases for expressions
```

**Fig. 2.** Anatomy of a refinement rule

```
a subseteq b ⇝ forall i : a . i elem b
    guards: a ∼ set of τ, b ∼ set of τ
```

**Fig. 3.** An example rewrite rule, ruleSetSubsetEq

`set`, `mset`, `function`, `relation`, `tuple`. Some of the available operators are the following: `abs`, `+`, `-`, `*`, `/`, `%` `=`, `!=`, `<`, `>`, `<=`, `>=`, `not`, `/\`, `\/`, `=>`, `<=>`, `card`, `elem`, `union`, `intersect`, `subset`, `subseteq`, `supset`, `supseteq`, `max`, `min`, `forall`, `exists`, `sum`, *matrix indexing* (`[]`), *tuple indexing* (`<>`), *function application*, *function inverse application*, *relation projection*.

It is useful to view our rules as operating upon an *Abstract Syntax Tree* (AST) representation of an ESSENCE specification. In the AST, every node represents a term in the specification and is also labelled with that term's type. The rewriting system works by traversing the AST and attempting to apply the rules in the database at every node. A rule is allowed to modify the subtree rooted at the current node, and, for contextual information, is allowed to access (but not to modify) the remainder of the AST via the parent of the current node. If a rule matches the current node, the whole subtree is replaced with the equivalent subtree the rule suggests.

### 4.1 Example Rules

In this section we will present a number of example rules relating to the refinement of set variables. These will demonstrate rules showing how expressions involving sets are rewritten and refined.

We first present a series of rules that demonstrate how expressions involving sets can be rewritten into expressions involving simpler operators.

```
a = b ⇝ a subseteq b /\ b subseteq a
    guards: a ∼ set of τ,  b ∼ set of τ
```

`ruleSetEq`: shows a rule that matches with an equality constraint between two sets. It rewrites the equality into a conjunction of mutual `subseteq`'s between the two sets. The guards ensure it only applies to sets.

```
e elem s ⇝ exists i : s . i = e
    guards: e ∼ τ, s ∼ set of τ
```

`ruleSetElem`: shows a rule that matches with an `elem` constraint and rewrites it into an existential quantification over the set.

```
forall i : (a union b) . k ⇝ forall i : a . k /\ forall i : b . k
exists i : (a union b) . k ⇝ exists i : a . k \/ exists i : b . k
forall i : (a intersect b) . k ⇝ forall i : a ( i elem b => k )
forall i : (a intersect b) . k ⇝ forall i : b ( i elem a => k )
exists i : (a intersect b) . k ⇝ exists i : a ( i elem b /\ k )
exists i : (a intersect b) . k ⇝ exists i : b ( i elem a /\ k )
    guards: a ∼ set of τ, b ∼ set of τ
```

`ruleSetQuan`: shows a series of rules that handle quantification expressions on some set expressions. These simple transformations save us from creating intermediate set variables.

The following series of rules show how, once constraints are reduced into a simpler set of operators, set variables are refined into lower-level types. CONJURE supports refining more complex expressions directly into constraints on lower-level types, but we have found this method of refinement reduces our rule set, without reducing the set of models we can generate.

```
quan i : s . k ⤳ quan i : r . kk
    guards: s ∼ set of τ, size(s) ∼ bound, refinement(s) ∼ explicit
            quan ∼ {forall,exists,sum}
    declarations: m  = matrix indexed by [r] of τ
                  r  = int(1..size(n))
                  kk = replace (i -> m[i]) k
```

`ruleRefineSetQuan (set size bound, explicit representation)`: shows one of the core rewrite rules for handling set refinements. This rule is not only a simple reformulation of an expression, but it also refines the set into an explicit representation, using a matrix. `refinement` is a function that returns the representation of its parameter. The `size` function, which returns the size of its parameter is used here to check if the set is of a fixed size.

```
forall i : s . k ⤳ forall i : r . kk
    guards: s ∼ set of τ, size(s) ∼ not bound, maxsize(s) ∼ bound
            refinement(s) ∼ explicit
    declarations: m  = matrix indexed by [r] of tuple<τ,bool>
                  r  = int(1..maxsize(n))
                  kk = (m[i]<1> = true) => replace (i -> m[i]<0>) k
exists i : s . k ⤳ exists i : r . kk
    guards: s ∼ set of τ, size(s) ∼ not bound, maxsize(s) ∼ bound
            refinement(s) ∼ explicit
    declarations: m  = matrix indexed by [r] of tuple<τ,bool>
                  r  = int(1..maxsize(n))
                  kk = (m[i]<1> = true) /\ replace (i -> m[i]<0>) k
sum i : s . k ⤳ sum i : r . kk
    guards: s ∼ set of tau, size(s) ∼ not bound, maxsize(s) ∼ bound
            refinement(s) ∼ explicit
    declarations: m  = matrix indexed by [r] of tuple<τ,bool>
                  r  = int(1..maxsize(n))
                  kk = m[i]<1> * replace (i -> m[i]<0>) k
```

`ruleRefineSetQuan (set size not bound, explicit representation)`: shows some extensions to the rule `ruleRefineSetQuan`. These rules handle cases where there is a quantified expression over a set whose size is not known in advance. An explicit representation of a set with unknown size requires introducing a matrix of tuples, where in each tuple the first component is in the set if the second component is true. For this representation we need different rules for each quantifier. `maxsize` is similar function to `size`.

```
forall i : s . k ⤳ forall i : r . kk
    guards: s ∼ set of int, refinement(s) ∼ occurrence
    declarations: m   = matrix indexed by [dom] of bool
                  dom = domain(tau(s))
                  kk  = (m[i] = true) => k
exists i : s . k ⤳ exists i : r . kk
    guards: s ∼ set of int, refinement(s) ∼ occurrence
    declarations: m   = matrix indexed by [dom] of bool
                  dom = domain(tau(s))
                  kk  = (m[i] = true) /\ k
sum i : s . k ⤳ sum i : r . kk
    guards: s ∼ set of int, refinement(s) ∼ occurrence
    declarations: m   = matrix indexed by [dom] of bool
                  dom = domain(tau(s))
                  kk  = (m[i] = true) * k
```

`ruleRefineSetQuan (occurrence representation)` : shows the final part of `ruleRefineSetQuan`. These rules handle cases where the set is to be represented in occurrence representation, where a `ith` element of a matrix is true if `i` is in the set. `domain` is a function which returns the domain of its parameter. `tau` is a function which returns the type of element a container type contains.

## 5 Matching Expressions not Constraints

The prototype implementation discussed in [4] operated by matching against and rewriting *complete constraints* after flattening all expressions by introducing auxiliary variables and further constraints. However, such an approach has a number of drawbacks. It may not be scalable in general as we may possibly have a huge number of constraint types that look very similar but slightly different (such as: `x subseteq (a union b)` and `x supseteq (a union b)`). Furthermore, a large number of rewrite rules may be needed, one for each constraint type. Finally, the flattening process may introduce a large number of unnecessary auxiliary variables and changes the structure of our constraints for which we may have better rewrite rules that exploit that structure.

In this paper, we overcome these drawbacks by allowing our rules to match and rewrite *expressions* within a constraint rather than (necessarily) the whole constraint. This allows us to accomplish three things. First, we can refine a greater proportion of the ESSENCE language using fewer rules. Second, unlike the prototype, we no longer need to *flatten* a specification prior to refinement avoiding introducing unnecessary auxiliary variables. Finally, we may have optimised rewrite rules for specific structured constraints. For instance, consider the constraint `(a union b) subseteq c`, if we flatten it, we would have `x subseteq c /\ x = a union b` which introduces an auxiliary variable and requires refining unnecessarily a set equality constraint. In our approach, in addition to this refinement, we may rewrite this into a conjunction of two `subseteq` constraints, namely `a subseteq c /\ b subseteq c`, by having a dedicated rewrite rule which reasons about the structure of this constraint type.

There is a subtle problem arising when we match an expression fragment and rewrite it to an equivalent expression fragment. The rewrite might introduce extra constraints and auxiliary variables.

For instance, consider the following Essence specification:

```
given lb,ub,n,m,k : int
find t : set (size n) of int(lb..ub)
find A : set (size n) set (size m) of int(lb..ub)

such that
 forall s : A . (max(s) - max(t) = k) => (k elem s)
```

The rewrite rule for the set max operator (`max(s)`) needs to introduce an auxiliary variable, say `max_s`, along with constraints that enforce that `max_s` is the maximum element in set `s`. We refer to these extra constraints as *helper constraints*.

We equip our rewrite rules with an extra operator "`@`" which attaches a "bubble" to any expression containing the helper constraints. For example, our rewrite rule for the set max operator is as follows:

```
max(s) ⇝ max_s @ bubble
    guards: s ~ set of int
    declarations: max_s  = int
                  bubble = (max_s elem s) /\ (forall i : s . i >= max_s)
```

If we apply our rule of the above example which contains two set max operator, we end up with the following resulting expression:

```
forall s: A . ((max_s@bubble_s) - max(t) = k) => (k elem s)
forall s: A . ((max_s@bubble_s) - (max_t@bubble_t) = k) => (k elem s)
```

As we can see, the intermediate expression is not a valid one yet. In fact, we need to move the bubbles to their correct positions. To achieve this, we introduce a rule which we call "`ruleBubbleUp`" which moves the bubbles one level up each time. This rule does not apply to *universally or existentially quantified* expressions. We will show later how we handle these cases.

For each applicable expression `exp`, the `ruleBubbleUp` rule goes through each child of that expression and if that child has a bubble attached to it, it moves the bubble one level up and attaches it to expression `exp`. If there is more than one child with a bubble, we attach them in the form of a conjunction to expression `exp`.

Without loss of generality and for the sake of simplicity, we give the definition of the `ruleBubbleUp` assuming that the expression has only child. In our current system, we handle expressions with any number of children.

```
exp ⇝ exp' @ b
    guards: exp is not a quantified expression, child(exp) ~ (a @ b)
    declarations: exp' = replace (a @ b) in exp with a
```

In our running example, these are the results of successively applying the `ruleBubbleUp` rule:

```
forall s: A . (((max_s-max_t) @ (bubble_s /\ bubble_t))=k) => (k elem s)
forall s: A . (((max_s-max_t=k) @ (bubble_s /\ bubble_t))) => (k elem s)
forall s: A . (((max_s-max_t=k) => (k elem s)) @ (bubble_s /\ bubble_t))
```

After finding the right positions for the bubbles, our next step safely converts "@" to conjunction "/\" resulting in the following valid expression:

```
forall s: A . (((max_s-max_t=k) => (k elem s)) /\ bubble_s /\ bubble_t)
```

Our optional last step is to resolve the scope for the bubble expressions inside a universally quantified expression. `ruleLoopInvariant`, defined below, moves all expressions inside a universally quantified expression that do not fall in the scope of their quantifier. The rule is defined as follows:

```
forall i :s . k ⤳ invariants /\ forall i : s . variants
   declarations: ks = splitConjunction k
                 variants  = filter (hasReferenceTo i) ks
                 invariants = ks \\ variants
```

Given a conjunction of constraints, the `splitConjunction` function returns a list of constraints breaking the conjunctions between them and the function `hasReferenceTo` checks whether the second expression references the first one or not. Thus, our final refinement is as follows:

```
bubble_t /\ forall s: A . (((max_s-max_t=k) => (k elem s)) /\ bubble_s)
```

## 6  Coverage of Essence and Limitations

The Essence language as defined in [4] contains five core type constructors: `set`, `multi-set`, `partition`, `tuple`, `function` and `relation`; and the simple types `int`, `bool` and user-defined enumerated and unnamed types. Currently our system has rules for handling specifications on all the type constructors, excepting `partition`. There are 22 operators on these type constructors defined in [4]. We support almost all of them. The currently missing operators are mostly related to type-conversion and we are confident that they will be easy to implement without changes to our current system. Since partitions can be realised as sets of sets with additional constraints, we are confident that we can support these in the near future.

We do not implement enumerated types and unnamed types yet. The initial focus of the implementation was attacking the challenging parts of Essence like the nested types, rather than providing a finished product. Since enumerated types and unnamed types can be easily transformed into `int` by a preprocessing step, we postpone the implementation of them.

```
not (a) ⤳ a = false
    guards: a ∼ bool
```

`ruleNot`: This rule removes negated expressions.

```
alldiff(m) ⤳ forall i : r . (forall j : r . (i < j) => m[i] != m[j])
    guards: tau(m) ≁ {bool,int}
    declarations: r = firstIndexOfMatrix m
```

**ruleComplexAlldiff**: We provide a generic implementation of `alldiff` which rewrites the constraint into a clique of not equal constraints, for use with types without a global *alldiff* constraint.

```
a != b          ⤳ not (a = b)           a supset b ⤳ b subset a
a supseteq b ⤳ b subseteq a             a subset b ⤳ a subseteq b /\ a != b
    guards: a ~ set of τ, b ~ set of τ
```

**ruleSetOps**: Shows a selection of the rules which handle normalising set expressions. This reduces the number of other rules which are required.

```
min(s) ⤳ min_s @ cons
    guards: s ~ set of int
    declarations: min_s = int
                  cons  = (min_s elem s) / (forall i : s . i >= min_s)
```

**ruleSetMin**: Rewrites a `min` operator applied to a set variable. `cons` constrains the newly introduced variable, `min_s`. `ruleSetMax` is defined similarly.

```
card(s) ⤳ size(s)
    guards: s ~ set of τ, size(s) ~ bound
card(s) ⤳ card_s @ (card_s = sum i : dom . m[i])
    guards: s ~ set of τ, size(s) ≁ bound, maxsize(s) ~ bound
            representation(s) ~ occurrence
    declarations: card_s = int
                  m     = matrix indexed by [dom] of bool
                  dom   = domain(tau(s))
card(s) ⤳ card_s @ (card_s = sum i : dom . m[i]<1>)
    guards: s ~ set of τ,  size(s) ≁ bound, maxsize(s) ~ bound
            representation(s) ~ explicit
    declarations: card_s = int
                  m = matrix indexed by [r] of tuple<τ,bool>
                  r = int(1..maxsize(n))
```

**ruleSetCard**: Rewrites a `card` operator applied to a set variable. Is optimised for varibles of known size, else it creates a variable to represent the cardinality of the operand set. There are variants for two representations of a set.

```
f(i) ⤳ m[i]
    guards: f ~ function τ₁ -> τ₂, i ~ τ₁
            representation(f) ~ Func1D
    declarations: m = matrix indexed by [dom(τ₁)] of τ₂
f(i) ⤳ sum j : r . j × m[i,j]
    guards: f ~ function τ₁ -> τ₂, i ~ τ₁
            representation(f) ~ Func2D
    declarations: m = matrix indexed by [dom(τ₁),dom(τ₂)] of bool
```

**ruleFuncApp**: Refines function application. for two different representations of functions

```
defined(f) ⤳ defn_f
    guards: f ∼ function τ₁ -> τ₂, isTotal(f) ∼ true
    declarations: defn_f = set of τ₁
defined(f) ⤳ defn_f @ bubbl
    guards: f ∼ function τ₁ -> τ₂, representation(f) ∼ Func2D
            isPartial(f) ∼ true
    declarations: defn_f = set of τ₁
                  m      = matrix indexed by [dom(τ₁),dom(τ₂)] of bool
                  bubbl  = forall i : dom(τ₂) .
                              ((sum j : dom(τ₁) . m[j,i]) > 0)
                              => i elem defn_f
```

**ruleFuncDefined**: Refines the `defined` operator applied to a function variable. Simply returns the domain set of the function, if the function is declared to be total, else it creates a set variable and constraints it.

```
range(f) ⤳ range_f
    guards: f ∼ function τ₁ -> τ₂, isSurjective(f) ∼ true
    declarations: range_f = set of τ₂
range(f) ⤳ range_f @ bubbl
    guards: f ∼ function τ₁ -> τ₂, representation(f) ∼ Func2D
            isSurjective(f) ≁ true
    declarations: range_f = set of τ₂
                  m      = matrix indexed by [dom(τ₁),dom(τ₂)] of bool
                  bubbl  = forall i : dom(τ₁) .
                              ((sum j : dom(τ₂) . m[i,j]) > 0 )
                              => i elem range_f
```

**ruleFuncRange**: Refines the `range` operator applied to a function variable. This rule is similar to **ruleFuncDefined**.

```
m[i]<j> ⤳ n<j>[i]
    guards: m ∼ matrix indexed by [indices] of tuple<components>
    declarations: n = tuple<comps'>
                  comps' = ∀ c ∈ components .
                              matrix indexed by [indices] of typeof(c)
```

**ruleMatrixOfTuples**: ESSENCE supports matrices of tuples, however the target language does not. This rule rewrites a matrix of tuples into a tuple of matrices, preserving the matrix indexing and the tuple indexing.

```
t<i> ⤳ tt
    guards: t ∼ tuple<components>, i ∼ integer expression
            length(components) >= i
    declarations: tt = typeof(components[i])
```

**ruleTupleOut**: Rewrites an indexed tuple into a separate decision variable.

```
a = b ⤳ forall i : r . a<i> = b<i>
    guards: a ∼ tuple<components₁>, b ∼ tuple<components₂>
            length(components₁) = length(components₂)
    declarations: r = int(0..length(components₁)-1)
```

`ruleTupleEq`: Rewrites tuple equality as equalities on the members of the tuple.

```
r<is> ⤳ is elem s
    guards: r ∼ relation of components₁, is ∼ tuple<components₂>
            length(components₁) = length(components₂)
            is contains no underscores
    declarations: s = set of tuple<components₁>
```

`ruleReElem`: Relation membership, representing the relation as a set of tuples.

```
r<is> ⤳ k<js> @ bubbl
    guards: r ∼ relation of components₁, is ∼ tuple<components₂>
            length(components₁) = length(components₂)
            is contains at least one underscore
    declarations: js = underscoredIndices(is)
                  k  = relation of js
                  bubbl = forall i : r .
                    (notUnderscoredIndices(is) match with i) =>
                    (underscoredIndices(i) elem k)
```

`ruleRelnProj`: Rewrites the relation projection expressions. In a relation projection expression, at least one of the tuple indices (`is`) should be left unspecified. This operator creates a new relation containing only those components of the actual relation which are left unspecified.

The only part of ESSENCE where we do not yet support full nesting is function variables that map from arbitrarily nested combinatorial objects to any type. We support only function variables that map from sets of integers to arbitrarily nested types. This is an important area of future work.

```
find f : function (partial) int(lb..ub) -> τ
find g : function (partial) set (maxsize n) of int(lb..ub) -> τ
```

For example, we can refine `f` to either a 1-dimensional matrix of decision variables of type $\tau$, or a 2-dimensional matrix of boolean decision variables, depending on the actual type of $\tau$. However we cannot refine `g` to a matrix in the same way because we cannot index a matrix with complex types.

The power of our system directly depends on the quality of our rules database. So far we have concentrated on achieving coverage of the ESSENCE language, rather than producing good models. A further future goal is to increase both the quantity and breadth of our rules in order to generate the widest possible set of models from a specification, in preparation for then selecting the best model or models from the set of models we produce.

## 7 Implementation

ESSENCE, like Zinc, is a domain-specific language (DSL) for writing abstract constraint specifications. Similarly, languages such as EaCL[13], ESRA[3], OPL[20], ESSENCE′ [17] and MiniZinc[15] are DSLs for writing constraint models. This is in contrast with constraint solvers such as Ilog Solver[8] and Gecode[19], which are implemented as a library in a programming language such as C++.

There are two standard methods for implementing DSLs. The first is to write a full parser and type checker for the language and map the language to an internal data structure that can be processed and executed. A second, simpler method is by extending an existing programming language. The DSL language is then called an embedded-DSL (EDSL).

We choose a hybrid approach. We implement the language as an EDSL to Haskell to leverage its power, yet we still implement a complete parser and a type checker that can read conventional ESSENCE specifications in. This may sound like duplicating some work, however, we make use of the EDSL while writing our rewrite rules. Using the EDSL for rewrite rules saves us a great deal of effort which would otherwise mean re-implementing features of the host language.

Some languages, like Haskell, are particularly suited to this task, and can be used to implement concise EDSLs that are easy to use. Haskell is a statically typed higher-order pure functional programming language. It provides a module system, an advanced type system with type classes and polymorphic types. It is particularly powerful in manipulating data structures using its powerful pattern matching infrastructure. These features make it a natural candidate for hosting EDSLs. There are EDSLs in Haskell for many fields including but not limited to 3D animations, image synthesis and production, and geometric region analysis. The `hmatrix-gplk` package [1] provides an EDSL for Linear Programming, as well as a solver.

Rhiger [18] provides further examples of EDSLs in Haskell, and describes the common techniques used to design them.

Building on top of Haskell gives us the ability to leverage Haskell's rich type system and grow from existing types and operations. As discussed in [5], ESSENCE itself is specified as a sequence of abstract lexemes. Our choice of concrete lexemes was made to suit the host language. The mapping to the abstract lexemes of the ESSENCE language definition is straightforward.

## 8 Conclusion

This paper has presented a new version of the CONJURE automated modelling system, which achieves far greater coverage of the ESSENCE language than previous versions. There remain a small number of areas of the ESSENCE language that we need to extend the system to cover, such as partition variables and function variables that map from arbitrarily nested types. This forms an immediate piece of future work. Further future work includes increasing the quantity and breadth of our refinement rules and beginning to select among the set of models produced.

**Acknowledgements** Ozgur Akgun is supported by a SICSA prize studentship. This research is supported by UK EPSRC grant no EP/H004092/1.

---

[1] http://hackage.haskell.org/package/hmatrix-glpk